\documentclass{article}

\usepackage[preprint,nonatbib]{neurips_2019}
\usepackage[utf8]{inputenc} 
\usepackage[T1]{fontenc}    
\usepackage{hyperref}       
\usepackage{url}            
\usepackage{booktabs}       
\usepackage{amsfonts}       
\usepackage{nicefrac}       
\usepackage{microtype}      
\usepackage{xspace}
\usepackage{wrapfig}
\usepackage{amsmath,amsfonts,latexsym,amssymb,amsbsy, amsthm,epsfig, theoremref, color, graphicx}
\usepackage{theoremref}
\usepackage{subcaption}
\usepackage{enumitem}
\usepackage{array}
\usepackage{authblk}
\newtheorem{theorem}{Theorem}
\newtheorem*{theorem*}{Theorem} 

\newtheorem{definition}{Definition}

\newcolumntype{M}[1]{>{\centering\arraybackslash}m{#1}}

\title{Robust Classification using Robust Feature Augmentation}
\newcommand{\swati}[1]{\textcolor{blue}{(Swati: #1)}}

\newif\ifsubmit
\submittrue

\ifsubmit
\newcommand{\kevin}[1]{}
\else
\newcommand{\kevin}[1]{\textcolor{cyan}{Kevin: #1}}
\fi

\newcommand{\ie}[0]{\emph{i.e.,}\xspace}
\newcommand{\etal}[0]{\emph{et al.}\xspace}
\newcommand{\eg}[0]{\emph{e.g.,}\xspace}

\begin{document}
 \author{Kevin Eykholt\thanks{University of Michigan, Ann Arbor, keykholt@umich.edu} ~~~Swati Gupta\thanks{Georgia Institute of Technology, swatig@gatech.edu} ~~~Atul Prakash\thanks{University of Michigan, Ann Arbor, aprakash@umich.edu} ~~~Amir Rahmati\thanks{Stony Brook University, amir@rahmati.com} ~~~Pratik Vaishnavi\thanks{Stony Brook University,  pvaishnavi@cs.stonybrook.edu} ~~~Haizhong Zheng\thanks{University of Michigan, Ann Arbor, hzzheng@umich.edu}}


\maketitle
\begin{abstract}
  
  Existing deep neural networks (DNNs), say for image classification, have been shown to be vulnerable to adversarial images that can cause a DNN  misclassification, without any perceptible change to an image. In this work, we propose  ``shock-absorbing" robust features such as {\em binarization} (\eg rounding) and {\em group extraction} (\eg color or shape) to augment the classification pipeline, resulting in more robust classifiers. Experimentally, we show that augmenting ML models with these techniques leads to improved overall robustness on adversarial inputs as well as significant improvements in training time. On the MNIST dataset,
  we achieved 14x speedup in training time to obtain 90\% adversarial accuracy compared to the state-of-the-art adversarial training method of Madry \etal \cite{madry2018towards}, as well as retained higher adversarial accuracy over a broader range of attacks. We also find robustness improvements on traffic sign classification using robust feature augmentation. Finally, we give theoretical insights for why one can expect robust feature augmentation to reduce adversarial input space. 
\end{abstract}

\section{Introduction}
Deep neural networks (DNNs) are used for various tasks, including image classification with
applications to character recognition, traffic sign classification, and autonomous driving. However, the pervasive use of DNNs has also raised concerns as to their robustness, and thus trustworthiness. Namely, existing DNNs have been shown to be vulnerable to {\em adversarial inputs}~\cite{szegedy2014intriguing}. These are inputs
that, to a human, appear similar to each other, but are assigned different labels by the DNN. 


Currently, there is an interest in designing networks that are robust to adversarial examples. Shafai \etal argue that adversarial robustness is limited based on the dimensionality of the input space \cite{shafahi2018are}.
Schmidt \etal suggest that accurate, but not robust models are a result of an insufficient number of training samples \cite{DBLP:journals/corr/abs-1804-11285}. Under a theoretical model in which it is possible to learn an accurate classifier from a single sample, they demonstrate that learning a robust classifier requires at least $O(\sqrt{d})$ samples. This problem manifests itself during training as the classifier learns to rely on predictive, but non-robust features. For example, Malhotra \etal added pixel noise to training inputs based on the true label of the input and found that the classifier learned to value the position of the noise pixel over any other feature when classifying the data~\cite{malhotra2019what}. Other works make similar findings, showing that traditional training of classifiers, results in a classifier learning highly predictive, but non-robust features and the classifier is thus exploitable \cite{2019arXiv190502175I,geirhos2018imagenettrained,2018robustness,2018explainability}. 

The main contribution of this paper is to propose a new approach, {\em robust feature augmentation}, as a component of standard machine learning techniques. In this approach, we augment a classification pipeline with {\em robust features} that are design to absorb most adversarial perturbations, thus improving the overall robustness of the classifier. Under a theoretical model, we provide results and characterizations that help explain as to why this approach improves robustness. Our work is also interesting in the light of recent works on certifiable robustness, for \eg Cohen \etal \cite{Cohen2019} mention that ``{\it it is typically impossible to tell whether
a prediction by an empirically robust classifier is truly robust to adversarial perturbations}'' however, with robust feature augmentation in the classification pipeline itself, one can expect robustness to bounded adversarial perturbations, by construction. 


Adversarial training, popularized by Madry \etal, is the current standard approach for designing robust machine learning models, in which L$_{\infty}$-bounded adversarial examples are generated during training. However, adversarial training is costly. As an alternative approach, we suggest augmenting a classification pipeline with robust features. Compared to Madry \etal~\cite{madry2018towards}, robust feature augmentation without adversarial training achieved 80\% adversarial accuracy 33x faster on MNIST. Additionally, combining robust feature augmentation with adversarial training achieved a 14x training time
speedup for achieving 90\% adversarial accuracy. Since robust feature augmentation works well with any DNN, we can get higher accuracy compared to recent attempts at certifiable robustness \cite{Raghunathan2018} on the MNIST dataset. 


A concurrently developed approach is to create a dataset that only contains robust features \cite{2019arXiv190502175I}. Previously, this approach was shown to improve the robustness of a trained model, but required precise manipulations of the dataset \cite{geirhos2018imagenettrained}. Using adversarial training on CIFAR, Ilyas \etal created an adversarially robust model, from which they identified robust and non-robust features. They removed the non-robust features from the dataset and showed that standard training on the robust dataset improved adversarial performance by about 45\% while decreasing test accuracy by 10\%. However, this approach improvement still failed to outperform the model adversarially trained on the original dataset. In this paper, we choose to improve robustness by identifying robust features that can be added directly to the classification pipeline, thus preserving standard training techniques. Our intuition is that since adversarial examples are a human-defined phenomenon, robust features can also be similarly defined.

\paragraph{Summary of contributions and outline of the paper:}  
\begin{itemize}
    \item We define the notion of a robust feature, computed from an input. Informally, a robust feature is a feature that does not change as input is adversarially perturbed. Typically, we intend these features to be meaningful attributes such as color and shape of an object being classified. But, it can also be a coarser categorization of the input that is expected to be stable under permitted adversarial perturbations (Section   \ref{sec:proposal}). 
    
    \item We show that a function computed on a set of robust features is also robust. In other words, we can use robust features an an input for robust classification decisions (Section \ref{sec:proposal}). 
    
    \item We show theoretical connections between (1) adversarially-trained classifiers that attempt to discover a non-linear separation boundary to maximize the separation between natural inputs and (2) using robust functions to map natural inputs to "pure" natural inputs and then using a linear classifier to separate the points (Section \ref{sec:proposal}). 
    
    \item  On MNIST, we use a binarization function as a robust feature and show that it improves the robustness of a standard classifier from 0\% to 74.64\% without any adversarial training. For an adversarially trained classifier, binarization reduces the training time by 14x for comparable adversarial accuracy and retains better accuracy as attack radius increases, e.g., 87.13\% adversarial accuracy vs 34.88\% for $\epsilon = 0.35$  as compared to ~\cite{madry2018towards} (Section \ref{sec:experiment}).
    
    \item  On a traffic sign dataset, we design a robust color extractor to augment a standard traffic sign classifier. Our augmented classifier prevents more than 90\% of adversarial attacks between signs of different colors (Section \ref{sec:experiment}).
\end{itemize}


Before delving into the details of the main contributions of our work, we give an overview of useful notation and definitions in Section \ref{sec:prelim}. 
\vspace{-0.3cm}
\section{Preliminaries}\label{sec:prelim}\vspace{-0.2cm}
In this section, we establish some notation and definitions that will be useful in the exposition of the remaining paper. We often refer to the set $\{1, \hdots, m\}$ as $[m]$ for the ease of notation. We assume there is an underlying data distribution $\mathcal{D}$ which the input set $\mathcal{X} \subseteq \mathbb{R}^n$ belongs to and each $x \in \mathcal{X}$ has a corresponding ground truth label $y \in [k]$. In particular, one can think of $\mathcal{X}$ as the set of inputs that a human (or an oracle) is able to classify. We follow the supervised machine learning setup where the basic goal is given a training dataset and corresponding labels $(x_i, y_i)_{i\in[N]}$, learn a function $F:\mathcal{X} \mapsto [k]$, that is a good approximation for the unknown function $f: \mathcal{X} \mapsto [k]$. We will assume that $f$ is such that $f(x_i) = y_i$ in the given data. More specifically, the goal is to seek to minimize the loss over a random sample over the input space $\mathbb{P}_{x \sim \mathcal{D}}(F(x) \neq f(x))$, which is often approximated by minimizing an empirical loss over a random training sample.  

It has been shown that although highly accurate approximations of $f(\cdot)$ can be learned, these approximations are not robust for with respect to perturbations of a majority of inputs. Let $d(\cdot,\cdot)$ be a distance function that measures the distance between inputs in $\mathbb{R}^n$ and let us denote an $\epsilon$-neighborhood of $x$, $B(x, \epsilon)$, as the set of points in $\mathcal{X}$ at a distance at most $\epsilon$ away from $x$, i.e., $B(x, \epsilon) = \{z \in \mathcal{X} ~|~ d(x,z) \leq \epsilon\}$ for some given $\epsilon > 0$. We call a function {\it robust} if it does not change its output over small neighborhoods around a subset of desired inputs $\mathcal{P} \subseteq X$. 

\begin{definition}
A function $F: \mathcal{X} \rightarrow [k]$ is said to be {\bf robust} over a subset $\mathcal{P} \subseteq \mathcal{X}$ with respect to $\epsilon>0$ if for all $x \in \mathcal{P}$: $F(x) = F(z)$ for all $z \in B(x, \epsilon)$. 
\end{definition}

We refer to $\mathcal{P}$ as ``pure'' inputs. Since the set of all possible inputs, $\mathcal{X}$, encountered in practice is assumed classifiable by a human (or an oracle), we can assume\footnote{Note that by this definition, the classifiable set of inputs is not convex since convex combination of two points in different neighborhoods may not lie in the neighborhood of any pure data point.} that $\mathcal{X} = \cup_{x \in \mathcal{P}} B(x,\delta)$ for some $\delta>0$. As an example, a constant function is robust over all inputs, by definition, however it may not be accurate. For some large enough $\epsilon$, the ground truth $f(\cdot)$ may itself not be robust, although it is accurate. Combining accuracy and robustness, we can define an adversarial input as follows:

\begin{definition} Given a ground truth function $f: \mathcal{X} \rightarrow [k]$ and a learned classification function $F: \mathcal{X} \rightarrow [k]$, suppose $F(x) = f(x) = f(z)$ for some $z \in B(x, \epsilon)$ and $F(x) \neq F(z)$, then $z$ is an {\bf adversarial example} for the classification function $F(\cdot)$.
\end{definition}

Suppose a function $F$ is robust\footnote{A related notion  is that of certifiable robustness that deals with the user being able to certify robustness of a given classifier, in the sense of property testing  \cite{Raghunathan2018}.} on a subset $R_{F, \epsilon} \subseteq \mathcal{X}$ with respect to $\epsilon$ (\ie $F(x) = F(z)$ for all $z \in B(x,\epsilon)$, $x \in R_{F, \epsilon}$). By  definition, any input in $R_{F,\epsilon}$ cannot be an adversarial example for the function $F$ with respect to $\epsilon$ and any arbitrary ground truth function $f$.


Ideally, we would like a robust classifier that is also {\bf accurate} on this input space, i.e., a classifier that minimizes the loss on pure inputs as well as the percentage of pure inputs that have adversarial inputs. Increases in robustness may result in a loss of accuracy, and the goal is be to find a feasible trade-off. For the rest of the paper, we will assume that $\epsilon$ is chosen small enough such that perturbing inputs in $\mathcal{P}$ within an $\epsilon$-neighborhoods does not change the ground truth classification, and we would like to compute classifiers that are robust over $\mathcal{P}$. 

\vspace{-0.3cm}
\section{Robust Feature Augmentation}\label{sec:proposal}\vspace{-0.2cm}
We propose two general techniques of developing robust classifiers: binarization (Section \ref{sec:binary}) and group feature extraction (Section \ref{sec:robust}). 
\begin{figure}[tb]
    \centering
    \begin{subfigure}[b]{0.2\textwidth}
        \includegraphics[height=3cm]{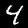}
        \caption{Original Image}
    \end{subfigure}
    \begin{subfigure}[b]{0.2\textwidth}
        \includegraphics[height=3cm]{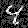}
        \caption{Adversarial image}
    \end{subfigure}
    \hspace{0.2cm}
    \begin{subfigure}[b]{0.2\textwidth}
        \includegraphics[height=3cm]{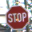}
        \caption{Original Image}
    \end{subfigure}
    \begin{subfigure}[b]{0.2\textwidth}
        \includegraphics[height=3cm]{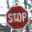}
        \caption{Adversarial image}
    \end{subfigure}
    \caption{\footnotesize The MNIST image in (a) is correctly classified as a ``4", however image in (b) is misclassified as an ``8", despite only minor visual distortions in the image. Similarly, the image in (c) is correctly classified as a {\sc Stop} sign, but the image in (d) is misclassified as German {\sc Keep Left} sign.}
    \vspace{-0.5cm}
    \label{fig:advexample} 
\end{figure}
First, we propose that if the first stage of a deep learning pipeline is robust to a class of perturbations, then the overall pipeline will also be robust against those perturbations. An example of binarization is a simple rounding filter that, when applied to an image, can remove perturbations on most pixels. Such a function is useful in images where there is a notion of a static background and only the presence of a single type of pixel defines the object. Previous works have demonstrated that, for MNIST, binarization is remarkably effective in improving adversarial robustness with respect to small pixel perturbations  \cite{DBLP:journals/corr/abs-1805-07816, Graese2016AssessingTO}. 
We will present theoretical reasons in Section \ref{sec:binary} on why the use of a binarizer improves robustness even without requiring adversarial training for the special case of a linear classifier, as well as present  experimental results on MNIST in Section \ref{sec:experiment} that show improved adversarial accuracy with this simple, yet powerful idea.  

Our second proposal is a generalization of binarization: to use one or more simpler image features (\eg color and shape for objects) that are expected to be robust to adversarial perturbations. Consider the domain of traffic sign images in the US: a standard {\sc Stop} traffic sign is known to be predominantly red and with octagonal shape. Traditional adversarial attacks on images change neither feature as there is a constraint to maintain the visual appearance of the original input (\eg  Figure \ref{fig:advexample}). Thus, it is apparent that standard classifiers do not learn to prioritize these features, shape and color, for labeling the sign. Rather, other predictive, non-robust features, are learned, which are then exploited by the adversary so as to maintain the visual appearance of the {\sc Stop} sign, while causing the predicted label to change. Our goal, then, is to make classifiers more robust by explicitly factoring in any known discriminating features that are robust to perturbations on a large subset of the input space.
\vspace{-0.2cm}
\subsection{Binarization}\label{sec:binary}\vspace{-0.2cm}
\begin{figure}[tb]
        \vspace{-0.2cm}
        \centering
        \includegraphics[width=0.9\linewidth]{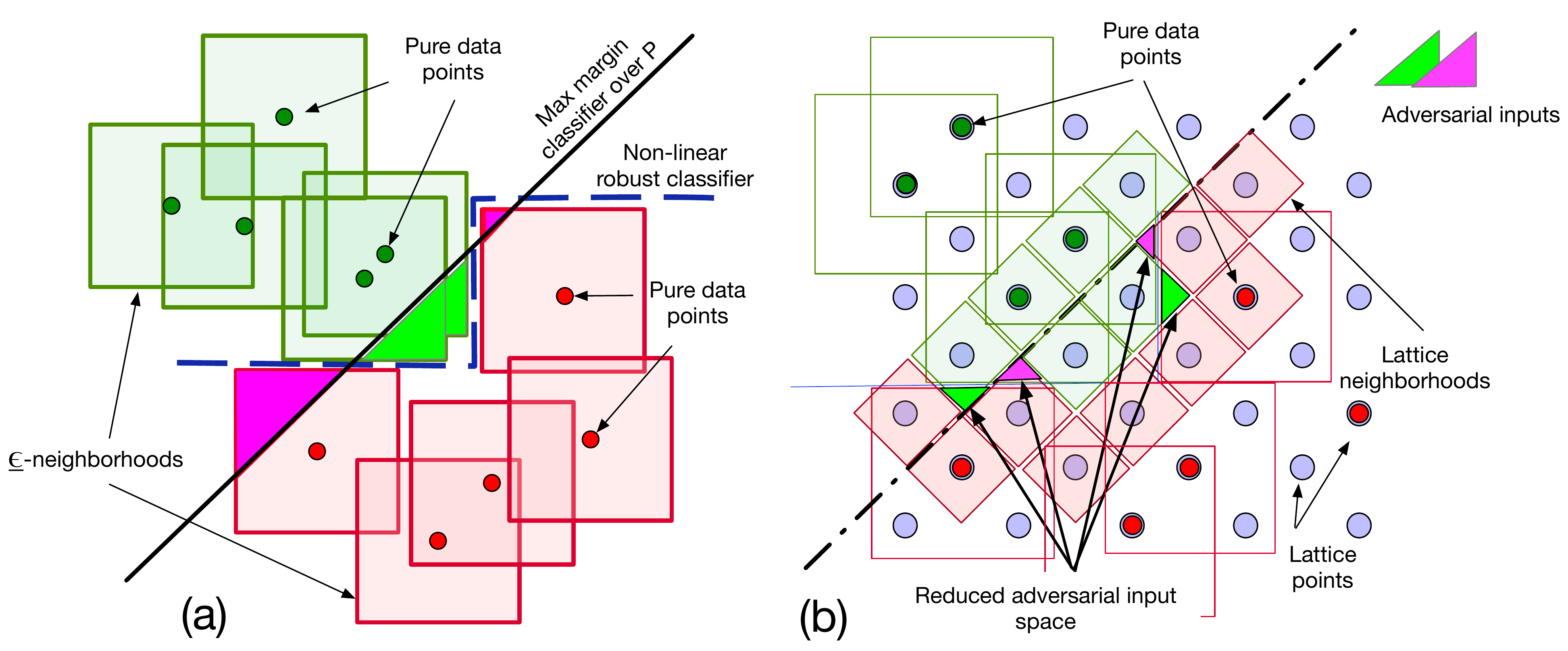}
        \caption{\footnotesize (a) Max-margin linear classifier, trained over pure data points $\mathcal{P}$, results in large adversarial input space. Binarizing test data to the nearest-neighbor in $\mathcal{P}$ before classification removes these adversarial inputs completely. (b) When $\mathcal{P}$ is not known, binarization to the nearest lattice point reduces adversarial input space.}
        \label{fig:combined}
        \vspace{-0.4cm}
\end{figure}
In order to remove spurious noise learned by a DNN, we propose {\it binarization} or a {\it snapping} of input data to desired intervals. Experimentally, we found that about 82\% of the pixels in MNIST images are concentrated near 0 and 8\% are concentrated near 1. The remaining pixels are somewhat evenly distributed between 0.1 and 0.9. We observed that adversarial attacks often changed background pixels, and if the changes were removed, the classifier would correctly label the example. Previous work suggests that a binarization function, which rounds all the pixels to \{0,1\} based on a threshold, can improve robustness of the resulting classifier \cite{DBLP:journals/corr/abs-1805-07816,Graese2016AssessingTO}. Although its name suggests rounding values in [0, 1] to \{0, 1\}, we define binarization more generally:

\begin{definition}
Consider a set $\mathcal{S} \subseteq \mathbb{R}^n$. Any function $b(\cdot)$ that maps each data point in  the input space $\mathcal{X} \subseteq \mathbb{R}^n$ to elements in $\mathcal{S}$ is called a binarizer, and $b(x)$ is referred to as the binarization of $x \in \mathcal{X}$.
\end{definition}

Typically, $\mathcal{S}$ is chosen to be much smaller in cardinality compared to $\mathcal{X}$. Suppose the binarizer $b$ is defined with respect to a distance $d(\cdot)$ such that $b(x)$ is the nearest neighbor of $x$ in $\mathcal{S}$, i.e. $b(x) \in \arg\min_{z \in \mathcal{S}} d(x,z)$. If $\mathcal{S} = \mathcal{P}$, we get a binarizer to map any data point to the nearest neighbor from the pure set of points $\mathcal{P}$. If $\mathcal{S} = \{0,1\}$ and $d(x,y) = \|x-y\|_{\infty}$ we get the vanilla form of binarization where every pixel is rounded to 0 or 1. 
One could also define a binarizer with respect to a threshold\footnote{Here, $\mathbb{I}_A$ is simply an indicator vector for whether $A$ is true or not.}, for e.g. $b(x) = \mathbb{I}_{x_i \geq \tau}$, which rounds each element to $\mathcal{S} = \{0,1\}$ based on whether the coordinate-wise value is less than threshold or not. 

We show in Section \ref{sec:experiment} that the proposed binarization has a minimal effect on the standard accuracy of the classifier, but greatly improves the adversarial robustness. Furthermore, binarization can be combined with adversarial training. The combination achieved both an order of magnitude faster training time and higher adversarial accuracy as compared to Madry \etal~\cite{madry2018towards}, with similar test accuracy. 


\vspace{-0.1in}

\paragraph{Why does binarization help?}

To see why binarization works in practice, consider the example of a support vector machine that computes a max-margin linear classifier. In Figure \ref{fig:combined}(a), suppose the set of ``pure" data points $\mathcal{P}$ are the green and red dots, and their $\epsilon$-neighborhoods in the $L_{\infty}$ norm are the colored squares enclosing them. In this example, the pure data points are linearly separable, although the $\epsilon$-neighborhoods are not.  
We depict the max-margin linear classifier with a solid line that separates the green points from the red points. Clearly, this example has a large adversarial instance space (pink, bright green regions in Figure \ref{fig:combined}(a)) which belongs to an $\epsilon-$neighborhood of some pure data point, however, these would be misclassified by the linear classifier. On the other hand, suppose a data point was first binarized to nearest-neighbor in $\mathcal{P}$, this would {\it completely remove} adversarial instances and we could obtain a perfect classifier even with the underlying classification technique being a support-vector machine. This point is important enough to be stated again: 

 \begin{centering}
 {\it Augmenting the classification pipeline with a nearest-neighbor mapping increases the power of linear classification to allow non-linear separability (blue decision boundary in Figure \ref{fig:combined}(a)). }
\end{centering} 

\begin{wrapfigure}{r}{0.4\textwidth}\vspace{-0.3cm}
    \centering
    \includegraphics[width=0.35\textwidth]{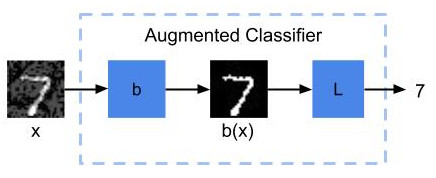}
    \caption{\footnotesize The MNIST model with a binarization function $b$ and classifier $L$.}
    \label{fig:binarized_mnist}
    \vspace{-0.2cm}
\end{wrapfigure}

Note that the resultant model from augmenting binarization and linear classification (see Figure \ref{fig:binarized_mnist}) is not only powerful in removing adversarial samples but also does so in an interpretable way. We formalize this example in the theorem below. 

\begin{theorem}
Consider a max-margin classifier that is trained on $\{(x_i, y_i)\}_{i=1}^N = \mathcal{P}$ that is linearly separable. Consider a distance function $d: \mathcal{P} \times \mathcal{P} \rightarrow \mathbb{R}$ and a parameter $\epsilon>0$. For any two data points $x_i, x_j \in \mathcal{P}$, suppose that the $d(x_i, x_j) > 2\epsilon$ whenever the ground truth labels $y_i \neq y_j$. Consider a nearest-neighbor binarizer $b(x) = \arg \min_{z \in \mathcal{P}} d(x,z)$, and the max-margin linear classifier $L$ (trained over $\mathcal{P}$), then the augmented classifier $C(x) = L(b(x))$ is {\bf robust} over $\mathcal{P}$ with respect to $\epsilon$ and exact\footnote{By {\it exact}, we mean no errors in classification.} over the $\epsilon-$neighborhood. 
\end{theorem}

The theorem holds because $b(x)$ is uniquely (and correctly) mapped to the original (unperturbed) data point using the nearest-neighbor binarizer, and these are perfectly classified using $C=L(b(\cdot))$ since $\mathcal{P}$ is linearly separable (therefore $L(\cdot)$ did not introduce errors on data points in $\mathcal{P}$). Since the $\epsilon$-neighborhoods of oppositely classified points do not overlap, we are able to perfectly classify the perturbed points using a linear classifier composed with nearest neighbor matching.

In the case when $\mathcal{P}$ is not known, we use binarization to map training/testing data points to the nearest points in a lattice, \eg the set of all 0/1 vectors $\{0,1\}^n$. This binarizer naturally acts as a regularizer for the output function since outputs in the neighborhoods of lattice points cannot change with small perturbations. Classification boundary over 0/1 vectors is much simpler than over the original (non-binarized) adversarial data. In our experiments, we augment a DNN with a lattice binarizer, which already gives compelling experimental results without adversarial training. We depict the reduction in adversarial input space in Figure \ref{fig:combined}(b). 
\vspace{-0.2cm}
\subsection{Group Feature Extraction}\label{sec:robust}\vspace{-0.2cm}

A generalization of binarization is to extend the notion to a collection of features (such as color, shape, size) that are found to be robust to perturbations. We think of a data point as a member in a group  defined by the value of such a feature (\eg {\sc STOP}, {\sc Do Not Enter} are members of the ``red'' color group). Given a predominantly red US traffic sign, it will require a large perturbation to change the majority of the sign to another sign color. However, unlike binarization to a lattice or $\mathcal{P}$, features like color or shape lie in a much smaller dimension, and lose the finer classification information. We propose two architectures for classification that can incorporate robust group features: (i) {\it intersection of multiple group features}, and (ii) {\it augmentation with original classifier}. 

\begin{wrapfigure}{r}{0.45\textwidth}
  \centering
  \vspace{-0.2cm}
  \begin{subfigure}{0.2\textwidth}
    \includegraphics[width=\textwidth]{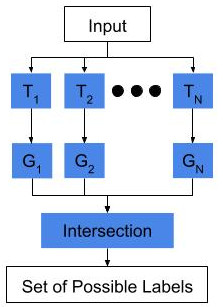}
     \caption{\footnotesize Multiple group feature extractors}
    \label{fig:parallel}
    \end{subfigure}
  \hspace{0.2cm}
    \begin{subfigure}{0.2\textwidth}
      \includegraphics[width=\textwidth]{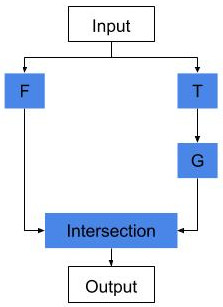}
      \caption{\footnotesize Augmented classifier design}
      \label{fig:augment}
  \end{subfigure}
\caption{\footnotesize Basic architecture of a robust classification network using group classifiers.}\vspace{-0.6cm}
\end{wrapfigure}

In the first architecture (Figure \ref{fig:parallel}), we propose to use multiple robust group feature 
 extractors $T_i$ each of which feeds the feature into $G_i$ to get a subset of possible labels. For example, suppose $T_1$ extracts the dominant sign color (\eg red) then $G_1$ can map the color to a set of possible road signs with the color (\eg map "red" color to \{{\sc Stop}, {\sc Do Not Enter}\}). This may not be enough information to get to the finer classification, however, adding another group extractor $T_2$ (\eg for shape) would allow one to classify more precisely (\eg identify {\sc Stop} or {\sc Do Not Enter}). The classifier output is simply the intersection of the possible labels given the extracted robust group features. We show that if all $T_i$ are robust, the resultant classifier formed by intersection is also robust. 
 \vspace{-0.1in}



\paragraph{Why do group features help improve robustness?} 
 Recall that in Section \ref{sec:prelim}, we defined $F: \mathcal{X} \rightarrow [k]$ as robust over a subset $\mathcal{P} \subseteq \mathcal{X}$ with respect to $\epsilon>0$ if for all $x \in \mathcal{P}$: $F(x) = F(z)$ for all $z \in B(x, \epsilon)$. For a given classification task that attempts to classify to labels in $[k]$, a {\it group feature} extractor can be viewed as a function $T: \mathcal{X} \rightarrow [m]$ that is robust with respect to $\gamma \gg \epsilon$ and maps to features in $[m]$ (typically, $m<k$). When referring to the architecture, we also refer to $T$ as a group feature extractor. The intuition here is that, if a group feature is known, then designing a feature extractor $T$, which is robust and accurate, is an easier task than learning a robust and accurate function $F$. Further, let $G: [m] \rightarrow 2^{[k]}$ map to possible labels given a group feature in $[m]$. We next show that the robustness guarantees naturally follow under function composition of $T$ and $G$:

\begin{theorem}
\thlabel{tm:2} Consider a group feature extractor $T: \mathcal{X} \rightarrow [k]$ that is robust on some subset of inputs $R$ with respect to $\gamma>0$, and a potential-label mapping $G: [m] \rightarrow 2^{[k]}$. Then the composition $G(T(\cdot)): \mathcal{X} \rightarrow 2^{[k]}$ is also robust on $R$ with respect to $\gamma$.
\end{theorem}
Theorem 2 holds since the internal group feature extractor $T$ acts as a {\it shock absorber} and the $G$ function is oblivious to the noise. Indeed for any $z \in B(x, \gamma)$ for $x \in R$, $T(x) = T(z)$ (due to robustness of $T$), and therefore, $G(T(x)) = G(T(z))$, i.e., $G(T.)$ is robust on $R$ with respect to $\gamma$. Robustness guarantees also hold in the case of intersection of multiple robust features:
\begin{theorem}
\thlabel{tm:3} Consider a set of robust feature extractors $T_i: \mathcal{X} \rightarrow [k]$ that are robust on some subset of inputs $R_i$ with respect to $\gamma_i$, and a sequence of potential-labels mappings $G_i: [m] \rightarrow 2^{[k]}$ for $i=1, \hdots, p$. Then the classifier that results by intersecting these: $C(x) =  \bigcap_{i=1}^p G_i(T_i(\cdot))$ is robust on $R = \bigcap_{i=1}^p R_i$ with respect to $\gamma = \min_{i=1, \hdots, p} \gamma_i$.
\end{theorem}
Theorem 3 holds trivially if $R = \emptyset$. Now consider $x \in R$ and $\gamma$ as defined. Then, for any $z \in B(x,\gamma)$, we have $G_i(T_i(z)) = G_i(T_i(x))$ using Theorem 2. Therefore, $C(z) = \cap_{i=1}^p G_i(T_i(z)) = \cap_{i=1}^p G_i(T_i(x))$ for all $z \in B(x, \gamma)$ and $x\in R$. 


One limitation of the above architecture is that we may not know sufficient robust features to make an unambiguous classification. To address this, we propose the {\em augmented architecture} (Figure \ref{fig:augment}). Specifically, we  deploy two networks in parallel, a group feature extractor of Figure \ref{fig:parallel} operating in parallel with a standard classifier $F$. The output of group-based network will be classification possibilities and we require outputs of $F$ and $G$ to be consistent with each other. This prevents targeted attacks on $x$ that change to a label $\notin G(T(x)$, e.g., changing a {\sc Stop} label (red) to a {\sc traffic light ahead} (yellow) label, thus reducing adversarial attack space.

This idea itself is quite powerful since it helps the DNN flag outputs where there might be an inconsistency: 
{\it Consider an augmented classifier $C(x) = F(x) \cap G(T(x))$. When $C(x) = \emptyset$ and $G(T(\cdot))$ is exact (\ie no errors), then we know that $F(x)$ was definitely an example that was misclassified.} This can be very useful in practice, where a machine can flag a difficult instance of data, and let an oracle (or a human) take over in these cases until $F(\cdot)$ can be made more accurate. We formalize this in Appendix \ref{app:theorem}. We show next that in some datasets, like US traffic signals, these ideas can help develop more robust classifiers.



\vspace{-0.3cm}\section{Experimental Results}\label{sec:experiment}\vspace{-0.2cm}
In this section, we present two sets of experiments demonstrating how our robust feature theory can be applied to improve classifier performance in an adversarial environment.

\paragraph{1. Binarization Augmentation on MNIST:} We start with a simple classification task, digit classification on the MNIST dataset~\cite{mnist}, and show that a binarization function both improves adversarial robustness and reduces training time compared to adversarial training to achieve a similar level of adversarial robustness. We use the pre-trained natural and adversarially trained MNIST classifiers used by Madry \etal \cite{madry2018towards}. For the attack, we use the PGD momentum attack code created by Zheng \etal \cite{DBLP:journals/corr/abs-1808-05537}. Our experiments compare four models, two of which use proposed binary augmentation:

\begin{enumerate}
    \item {\bf Natural Model {\sc (Natural)}:} Madry \etal's pre-trained natural classifier.
    \item {\bf Madry \etal's Adv. Trained Model {\sc (MAT)}:} Madry \etal's pre-trained robust classifier.
    \item {\bf Binarized Natural Model {\sc (BIN)}:} A natural classifier with a binarization function as the first processing step, trained on the natural training data (no adversarial training).
    \item {\bf Binarized Adv. Trained Model {\sc (BAT)}:} A classifier with a binarization function at the input, with the overall classifier trained on  adversarially perturbed training data.
\end{enumerate}

\begin{wraptable}{r}{2.5in}
\vspace{-0.2cm}
\centering \footnotesize  
\caption{\footnotesize The accuracy of each model evaluated against the MNIST test set and L$_{\infty}$ perturbations within $\epsilon=0.3$.}
\label{tab:accuracy}
\begin{tabular}{|p{0.66in}|p{0.66in}|p{0.66in}|}
\hline
Model & Test Acc. & Adv. Acc. \\
\hline
{\sc Natural} &	99.17\% & 0\%  \\ \hline
{\sc BIN*} &	98.93\% & \textbf{74.64\%} \\ \hline 
{\sc MAT} & 98.04\% & 89.72\% \\ \hline
{\sc BAT*} & 99.29\% & \textbf{91\%} \\
\hline
\end{tabular}
\end{wraptable} 

All models use the same model architecture (same as used in \cite{madry2018towards}). {\sc BIN} and {\sc BAT} include a binarization function, encoded as a step function centered at a threshold $\tau$, at the input of the network. Any pixel which is below $\tau$ ($\tau=0.5$ by default) is set to 0; else it is set to 1. For {\sc BAT} and {\sc MAT}, we generated adversarial examples in $B(x,0.3)$ for any given $x$, we run 100 iterations of the PGD attack with a step size of 0.0075 and 20 random restarts. As in the original experiments done by Madry \etal~\cite{madry2018towards}, an adversarial attack on a particular input sample is considered successful if at least one of the 20 generated adversarial perturbations is successful in changing the predicted label. For BAT, since the step function is non-differentiable, we use the Backward Pass Differential Approximation (BPDA) technique to generate good adversarial examples, as suggested by Athalye \etal~\cite{DBLP:journals/corr/abs-1802-00420}.

We first evaluated the test and adversarial accuracy of all 4 models for $\epsilon=0.3$ (\ie using PGD to find adversarial examples for an input $x$ within $B(x, 0.3)$, see Table \ref{tab:accuracy}). We observe that  binarization greatly improves the adversarial accuracy of {\sc Natural} from 0\% to 74.64\% despite no adversarial examples being used during training. We see that {\sc BAT}, the binarized implementation of {\sc MAT}, improved adversarial accuracy from 89.72\% to 91.14\%. Test accuracy was over 98\% for all models.

\begin{figure}
\vspace{-0.2cm}
\begin{subfigure}[t]{0.5\textwidth}\centering
    \includegraphics[width=0.8\textwidth]{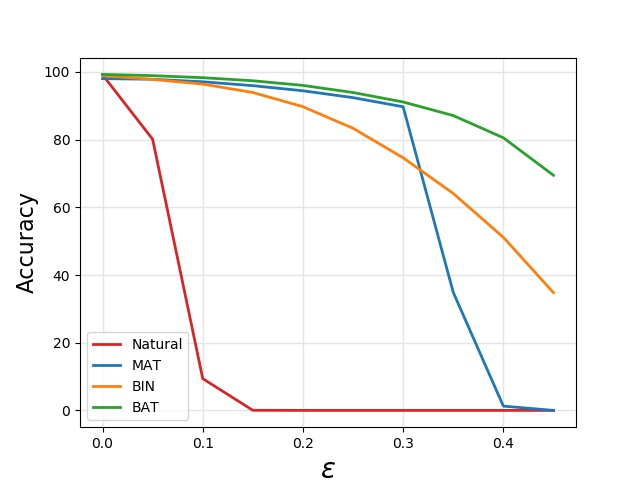}
\end{subfigure}
\begin{subfigure}[t]{0.5\textwidth}
 \centering
    \includegraphics[width=0.8\textwidth]{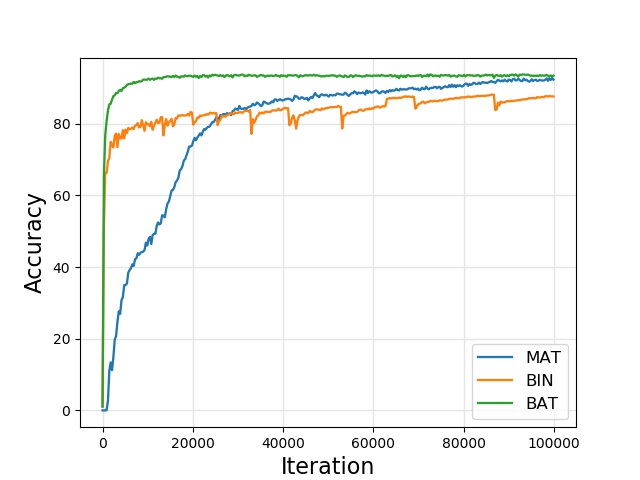}
    ~ 
\end{subfigure}

\caption{\footnotesize The adversarial performance during testing (left) and training (right). Not shown in the figure: {\em {\sc MAT} and {\sc BAT} take approximately 10x more time per training iteration than {\sc BIN}.}}\label{fig:vary}
\vspace{-0.4cm}
\end{figure}

We next measured the adversarial accuracy of all four models for different values of $\epsilon$ between 0 and 0.5. We emphasize that {\sc MAT} and {\sc BAT} are still trained for $\epsilon=0.3$; only the attacker's capabilities are changed. 
In Figure \ref{fig:vary}, we see that binarization is likely reducing attack space for large $\epsilon$ (e.g., {\sc BIN} outperforms {\sc MAT} when $\epsilon =  0.35$ with adversarial accuracy of 64.11\% versus 34.88\%, respectively). Also, adversarial training used with binarization further improves the robustness of the classifier (e.g., {\sc BAT} has an adversarial accuracy of 87.13\% for $\epsilon = 0.35$, more than double that of {\sc MAT}). 

The above findings can be particularly important in settings where adversarial training is infeasible, say for learning on edge computing devices with smaller computational budget.  {\sc BIN} itself, with no adversarial training, results in a significant  initial adversarial robustness. In {\sc MAT} and {\sc BAT}, each iteration of training is much more expensive since a PGD attack is executed to create a set of adversarial training examples. To further analyze the training efficiency, we evaluated the adversarial accuracy every 300 training iterations for both binarized models and {\sc MAT}\footnote{All training was done on a 12GB Titan X Pascal GPU}. Adversarial examples with $\epsilon=0.3$ were generated using 100 iterations of the PGD attack with a step size of 0.0075 and no random restarts\footnote{40 iterations with a step size of 0.01 is about twice as fast, but the adversarial accuracy of the model suffers}. These results are shown in Figure \ref{fig:vary}. We observe that although {\sc MAT} starts achieving a higher adversarial accuracy  than {\sc BIN} after about 30,000 iterations, each training iteration for {\sc MAT} took 236 ms versus 22 ms for {\sc BIN}. As a result, {\sc BIN} achieved 80\% adversarial accuracy after about 2.9 minutes of training versus 96 minutes of training for {\sc MAT}. In {\sc BAT}, where binarization is used during adversarial training, we see large reductions in training time required for comparable adversarial accuracy. {\sc BAT} achieved 80\% adversarial accuracy in about 3.6 minutes and 90\% accuracy in about 19 minutes. {\sc MAT} only achieved 90\% after 273 minutes of training (14x slower than {\sc BAT}).
\vspace{-0.1in}
\paragraph{2. Group Feature Extraction:}
\label{sec:results_signs}
We now move to a more complex task, traffic sign classification, and demonstrate how a using a robust function to extract a robust feature, the dominant color of a sign, can help reduce the adversarial attack space, e.g., preventing attacks that would change a classification across colors (e.g., red {\sc Stop} to a blue {\sc minimum speed 30} sign in Germany).

Based on the architecture shown in Figure \ref{fig:augment}, we augment a traffic sign classifier with a robust feature extraction pipeline, responsible for determining the dominant color of the sign and mapping the color to a set of possible traffic signs. Simply described, the color extractor first determines the sign's position in the image. Once located, it assigns each pixel a label based on the closest color center in the hue color space, either ``red", ``blue", or ``yellow", then outputs the color based on a weighted majority vote. A more detailed description of the extractor can be found in the appendix.

\begin{wraptable}{r}{0.45\textwidth}
\centering
\footnotesize
 \caption{\footnotesize \# Adv. image is the number of adversarial images ($\epsilon=8$) in which the predicted label matched the adversarial target. The correction rate is the percentage of adversarial examples for which the color extractor outputs red.}
 \label{tab:sign_results}
\begin{tabular}{p{28mm}M{10mm}M{10mm}}
\toprule
Adversarial Target & \# Adv. Images   &  Correction Rate \\
\toprule
Blue Signs (GTSRB)          & 13633  
& 93.53\%\\
Yellow Signs (LISA)              & 2389 
& 95.33\%\\
\bottomrule
Total \# of Stop signs & 3021 & \\
\bottomrule
\end{tabular}
\vspace{-0.1cm}
\end{wraptable}

We train a traffic sign classifier on a dataset composed of traffic sign images from both the LISA~\cite{lisa,lisadataset} and GTSRB sign dataset \cite{gstrbdataseturl,stallkamp2011a} (normal training). 
We then perform 20 iterations of a targeted L$_{\infty}$-bounded PGD attack with $\epsilon=8$ and step size of 2. The goal is to perturb a {\sc Stop} sign into a target sign class that is either blue or yellow. The performance is evaluated on 9 target sign class (8 blue sign classes, 1 yellow sign class) and reported in Table \ref{tab:sign_results}. 

Overall, the color extractor prevents over 93\% of above adversarial attacks that change {\sc Stop} to a blue or yellow sign (Table~\ref{tab:sign_results}). 
Of course, an attacker could attempt to adversarially attack the color extractor's robustness assumption.  Using the same set of {\sc Stop} sign images, we explored the edges of the $\epsilon$-neighborhood ($\epsilon=8$) for each image and checked if the color extractor's output changed at any point. From this, we found that the extractor is robust on approximately 75\% of the {\sc Stop} sign images. The ones that are not robust were poorer quality images, e.g.,  very dark images, a potentially preventable problem by requiring better lighting and using better cameras. We include robustness measurements for attacks on color for different values of $\epsilon$  in Appendix B.

\paragraph{Conclusion}
The existence of adversarial examples is attributed to a network's reliance on predictive, but easily exploitable, features it learned during training. In this work, we introduced two methods of \emph{robust feature augmentation} to mitigate this problem: {\em binarizers} and {\em robust group features}. Both map the input space to a smaller, more robust, subspace (like a lattice or group labels) and we formally describe these two methods to improve DNN robustness. Experimentally, we demonstrated how these methods can improve the adversarial robustness of a digit classifier and a traffic sign classifier. Furthermore, when adversarial training is used in conjunction with these methods, we were able to train a more adversarially robust model for MNIST 14x faster than without these methods.

We recognize that human identification of robust features may not be applicable to all machine learning tasks, especially if non-interpretable, robust features exist. As such, it is important to develop techniques to identify such features, though doing so is beyond the scope of this work. However, concurrent work done by Ilyas \etal has already shown some progress in this area, through the use of adversarial training to remove non-robust features from training data \cite{2019arXiv190502175I}. We expect to see further research, in which robust feature augmentation can be {\it the} method for adversarial robustness.

\section*{Acknowledgments}
We thank Yashu Liu of Didi Labs for his helpful feedback on the earlier draft of the paper. This project is supported by Didi Chuxing. This material is based in part upon work supported by the National Science Foundation under Grant No. 1646392.

\begingroup
\let\clearpage\relax
\onecolumn{\bibliography{ref}

\footnotesize}
\endgroup

\newpage
\appendix
\section{Additional Background}
\paragraph{Adversarial Training} 

Madry et al.~\cite{madry2018towards} proposed the use of adversarial training in which they solve 

\begin{equation*}
    \min_{\theta}{\rho(\theta)}, \; \text{where:} \; \rho(\theta) = \mathbb{E}_{(x,y) \sim \mathcal{D}} \big[\max_{\delta \in \mathcal{S}} L(\theta, x + \delta , y)\big]
\end{equation*}

In their formulation, $\mathcal{S}$ is the set of allowed perturbations. The loss function $L$ quantifies the loss relative to the perturbed input $x + \delta$ and the original label $y$. The inner maximization problem seeks to find a perturbation $\delta$ that maximizes the loss for a given input $x$. The outer minimization problem aims to find the model parameters $\theta$ such that the expected adversarial loss in the inner maximization problem is minimized.

\paragraph{Projected Gradient Descent (PGD)} 
Adversarial training of a model on input $x$ requires generating an adversarial example and then training the model on the adversarial example. Madry \etal use the PGD attack to generate adversarial examples \cite{madry2018towards}. The PGD attack is an iterative method in which at each step $t$ the input $x^t$ is modified based on the negative gradient of the loss function:
\begin{equation*}
    x^{t+1} = \mbox{II}_{x+S}(x^{t}+\alpha \mbox{ sgn}(\nabla_{x}L(\theta,x^t,y)))
\end{equation*}

$\mathcal{S}$ is the set of allowed perturbations as defined previously. $\mbox{II}_{x+S}$ is a clip function, which ensures the perturbed input $x^{t+1}$ is within the allowable range.

\section{Traffic Experiment Details}
\subsection{Traffic Sign Dataset}

Traffic signs are fairly standard across counties (e.g., see \url{https://www.autoeurope.com/roadsigns/} for classes of traffic signs and examples). LISA~\cite{lisadataset,lisa} and German Traffic Sign Recognition Benchmark (GTSRB~\cite{gstrbdataseturl,stallkamp2011a}) are two popular traffic sign datasets that have been extensively used in previous studies.

 We created a traffic sign dataset using images from both the LISA traffic sign dataset~\cite{lisadataset,lisa} and the German Traffic Sign Recognition Benchmark (GTSRB~\cite{gstrbdataseturl,stallkamp2011a}). The LISA dataset contains images of 47 different U.S. traffic signs. However, there are large class imbalances (\eg {\sc Stop} has 1821 images and {\sc Speed Limit 55} has 2 images). To address this problem, we first combine the LISA training and GTSRB training dataset, which has images for 43 German traffic signs classes. The image labelled as {\sc Stop} in both datasets are combined as they have the same visual appearance. Similarly, the images labelled as {\sc Do Not Enter} and {\sc StreetClosedOneWay} are combined.
 
The combined dataset still has low representation for some of the individual U.S. traffic signs. To address that, we created two superclasses composed of white rectangular U.S. traffic signs and yellow U.S. traffic signs. The first superclass  contains U.S. Speed Limit signs and {\sc Right Lane Must Turn}. The second superclass cotnains U.S. Warning signs and {\sc School}, which are yellow. The 45 class labels in the augmented dataset are provided in Table \ref{tab:dataset_decription}.

\begin{figure}[h]
    \centering
    \begin{subfigure}[b]{0.9\textwidth}
        \includegraphics[width=\textwidth]{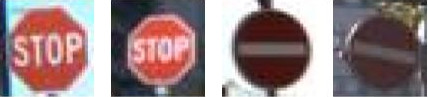}
        \caption{Red sign examples}
        \label{fig:red_examples}
    \end{subfigure}
    ~
    ~
    ~
    ~
    \begin{subfigure}[b]{0.9\textwidth}
        \includegraphics[width=\textwidth]{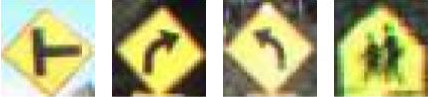}
        \caption{Yellow sign examples}
        \label{fig:yellow_examples}
    \end{subfigure}
    ~
    ~
    ~
    ~
    \begin{subfigure}[b]{0.9\textwidth}
        \includegraphics[width=\textwidth]{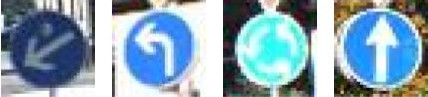}
        \caption{Blue sign examples}
        \label{fig:blue_examples}
    \end{subfigure}
    
    \caption{Examples images of signs for the three color classes we evaluated.}
    \label{fig:color_examples}

\end{figure}

With respect to the color extractor, we focused on signs of one of three dominant colors: red, yellow, and blue. U.S. red signs are generally regulatory in nature (e.g., {\sc Stop}, {\sc DoNotEnter}) and some examples are shown in in Figure~\ref{fig:red_examples}. U.S. yellow signs (see Figure~\ref{fig:yellow_examples}) are used  for cautioning a user (e.g., {\sc IntersectionAhead}, {\sc CurveRight}, {\sc CurveLeft}, {\sc School Zone}). Blue signs (see Figure~\ref{fig:blue_examples}) are common in Germany and can be restrictive or mandatory (e.g., {\sc KeepLeft}, {\sc MandatoryLeftTurn}, {\sc TrafficCircle} , {\sc MandatoryAhead}).  Table \ref{tab:color_split} identifies the sign labels in the modified dataset and that are either red and blue. For the purpose of classification, yellow signs are grouped together in a single class due to low representation with respect to the original sign labels(\eg {\sc Intersection}: 13 images, {\sc CurveLeft}: 24 images, {\sc TurnRight}: 24 images).

\begin{table}[h]
\centering
 \caption{\footnotesize Class labels of the LISA-GTSRB traffic sign dataset used in the experiments.}
\footnotesize
\begin{tabular}{M{33mm}M{33mm}M{33mm}}
\toprule
Class Label & Class Label  & Class Label \\
\toprule
speedLimit20 & streetClosedBothWays & wildlifeWarning \\
speedLimit30 & noTrucks & allRestrictionsEnd \\
speedLimit50 & generalWarning & mandatoryRightTurn \\
speedLimit60 & sharpLeftTurnAhead & mandatoryLeftTurn \\
speedLimit70 & sharpRightTurnAhead & mandatoryAhead \\
speedLimit80 & sequenceSharpTurnsAhead & mandatoryAheadOrRight \\
endSpeedLimit80 & bumpsInRoad & mandatoryAheadOrLeft \\
speedLimit100 & slipperyRoad & keepRight \\
speedLimit120 & tighterRoadOnRight & keepLeft \\
noPassing & construction & trafficCircle \\
noPassingTrucks & trafficLight & endNoPassing \\
intersectionWarning & pedestrianCrossing & endNoPassingTrucks \\
rightOfWay & schoolCrossing & Yellow Signs \\
yield & bicycles & doNotEnter \\
stop & icyRoads & White Rectangles \\
\bottomrule
Total \# of Signs & 44121 & \\
\bottomrule
\end{tabular}
 \label{tab:dataset_decription}
 \end{table}
 
 \begin{table}[h]
\centering
 \caption{\footnotesize Red and blue sign groupings. Yellow is not included as they have been grouped into a single label with respect to classification.}
\footnotesize
\begin{tabular}{M{33mm}M{33mm}M{33mm}}
\toprule
Red & Blue   \\
\toprule
Stop &  mandatoryRightTurn  \\
Do Not Enter & mandatoryLeftTurn  \\
& mandatoryAhead \\
& mandatoryAheadOrRight \\
& mandatoryAheadOrLeft \\
& keepRight \\
& keepLeft \\
& TrafficCircle \\
\bottomrule
\end{tabular}
 \label{tab:color_split}
 \end{table}

\subsection{Model Description}
We use a publicly available implementation of a multi-scale DNN architecture \cite{yadav}. The architecture description is given in Table \ref{tab:model-arch}. Before training, we triple the size of any class with less than 200 images through oversampling and random perturbations of each image. We use K-fold cross-validation with 10 splits and for each split, we train the model 50 times over the entire training split. Our trained model has 97.51\% test accuracy based on the GTSRB test dataset containing 12630 images. Our model was not adversarially trained, but the augmented pipeline does allow for an adversarially trained classifier, which may further improve robustness.

\begin{table}[h]
\center
\caption{Traffic sign model architecture. The model expects $32 \times 32 \times 3$ images as input with values in the range [-0.5, 0.5]. } 
\begin{tabular}{c c c c c c} \toprule
Layer Type & Number of Channels & Filter Size  & Stride & Activation \\ \toprule
conv & 3 & 1x1  & 1 & ReLU \\
conv & 32 & 5x5  & 1 & ReLU \\
conv & 32 & 5x5  & 1 & ReLU \\
maxpool & 32 & 2x2  & 2 & - \\
conv & 64 & 5x5  & 1 & ReLU \\
conv & 64 & 5x5  & 1 & ReLU \\
maxpool & 64 & 2x2  & 2 & - \\
conv & 128 & 5x5  & 1 & ReLU \\
conv & 128 & 5x5  & 1 & ReLU \\
maxpool & 128 & 2x2  & 2 & - \\
FC & 1024 & -  & - & ReLU \\
FC & 1024 & -  & - & ReLU \\
FC & 43 & -  & - & Softmax \\
\bottomrule \\
\end{tabular}
\label{tab:model-arch}
\end{table}

\newpage
\subsection{Color Extraction Algorithm}
\begin{figure}
    \centering
    \includegraphics[width=0.9\textwidth]{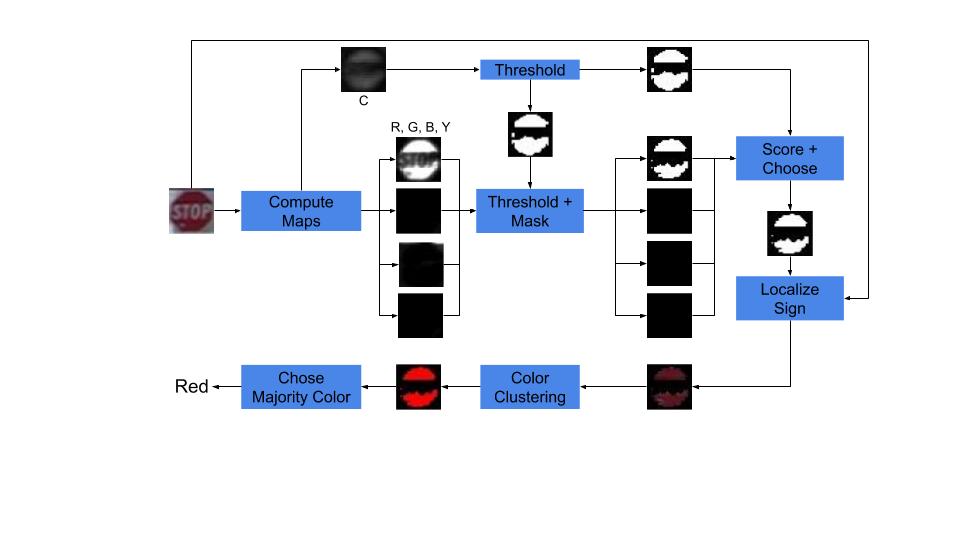}
    ~ 
      \vspace{-0.7cm}
    \caption{The color extractor pipeline. We show the step-by-step process for a {\sc Stop} image.}
    \label{fig:color_pipeline}
\end{figure}

We designed a basic color extractor for traffic sign classification. The extractor involves 2 steps:
\begin{enumerate}
    \item Sign Localization - Determine the sign's location in the image
    \item Color Classification - Determine the dominant color of the sign
\end{enumerate}
The full pipeline is shown in Figure \ref{fig:color_pipeline}.

\subsubsection{Sign Localization}
\label{sec:signlocalization}
Before we can evaluate the dominant color of the sign, we must first identify which pixels in the image correspond to the surface of the sign. Due to the presence of noisy image, like those shown in Figure \ref{fig:nonrobust_examples}, edge detection and contour extraction algorithms perform poorly. Instead, given a three channel color image, (r,g,b), we normalize each individual channel by the image intensity and compute a chromaticity map (C) and 4 color maps (R, G, B, Y)  \cite{channels, colorextract}.  
\begin{flalign}
\begin{split}
 & C = \max(r,g,b) - \min (r,g,b)\\
 & R = r - \frac{g+b}{2}\\
 & G = g - \frac{r+b}{2}\\
 & B = b - \frac{r+g}{2}\\
 & Y = \frac{r+g}{2} - \frac{|r-g|}{2} + b\\
 \end{split} \nonumber
\end{flalign}

Afterwards, all of the maps are converted to a binary image based on the mean of the non-zero values in each map. We use the binary image of C as a mask on each of the binarized color maps to  isolate the chromatic colors in each map. Finally, each channel is scored based on the number of non-zero pixels in the image. If less than 10\% of the pixels in each of the four color channels are white, the inverted binary chromaticity map is output. Otherwise, the binarized color channel with the highest score is output. We make one optimization based on the fact that in most of the images, the traffic sign is centered in the image. As such, we restrict thresholding and scoring to a small box around the center of the image. In our experiments, we used a 10 by 10 box.

\subsubsection{Color Classification}
The output of the sign localization step is to mask the original color image, and remove background pixels during color extraction. The image is converted to a hue-based representation (\eg HSV or HSL). Three predefined color centers (red, yellow, and blue) are used to label each non-zero pixel in the masked image based on the closest color center. Afterwards, a weighed weighted majority vote is computed (\ie weight of a pixel's vote increases the closer it is to the center) and the color with the most votes chosen.

For these proof-of-concept experiments, we choose to only detect red, yellow, and blue as these are the three most common colors in the dataset. We did not handle  colors such as brown or green as there were no signs in the dataset with these colors. Traffic signs that are white do exist, but white is not characterized by hue, but is instead based on the values of the other channels. As such, the color extractor is not robust for predominantly white signs, thus our analysis did not focus on such signs. This does not hurt the test accuracy of the augmented model, though, as we can include ``white" sign labels in the group-labels for all three colors. When we augment the classifier with the color extractor, the test accuracy on the GTSRB test dataset is still 97.51\%. Extending the color extractor to extract other colors, or even multiple colors, for finer-grain color-based classification, is future work.


\subsection{Robustness of the Color Extractor}
\begin{figure}
    \centering
    \includegraphics[width=0.5\textwidth]{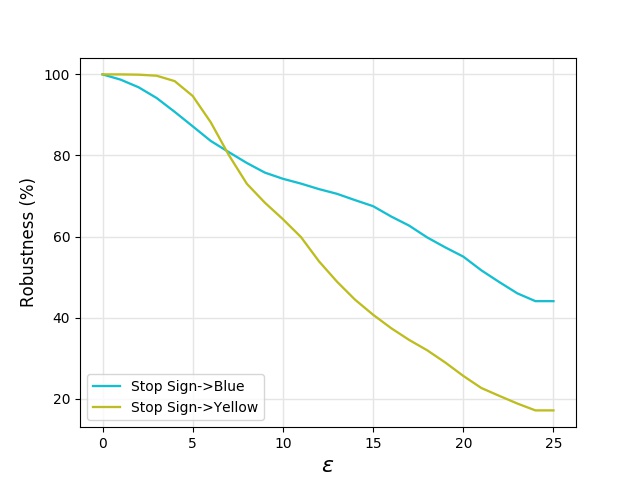}
    ~ 
    \caption{The robustness of the color classifier for {\sc STOP} when changing to blue or yellow signs as L$_{\infty}$ bound increases.}
    \label{fig:color_robustness}

\end{figure}

\begin{figure}
    \centering
    \begin{subfigure}[b]{0.17\textwidth}
        \includegraphics[width=\textwidth]{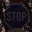}
        \caption{Close to Blue}
                \label{fig:bluebad1}
    \end{subfigure}
    ~
    ~
    ~
    ~
    \begin{subfigure}[b]{0.17\textwidth}
        \includegraphics[width=\textwidth]{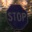}
        \caption{Close to Blue}
        \label{fig:bluebad2}
    \end{subfigure}
    ~
    ~
    ~
    ~
    \begin{subfigure}[b]{0.17\textwidth}
        \includegraphics[width=\textwidth]{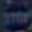}
        \caption{Close to Yellow}
        \label{fig:yellowbad}
    \end{subfigure}
    
    \caption{Some examples of inputs the color classifier is not robust on. Often, this occurs due to either the image being too dark (which tends to shift colors to blue) or the image being too blurry (which causes errors during sign localization).}
    \label{fig:nonrobust_examples}

\end{figure}

In Section \ref{sec:results_signs}, we presented the results on the robustness of the color extractor on {\sc STOP} images for an $L_{\infty}$-bounded attack with $\epsilon=8$. The evaluation involved shifting one or more color channels by $\epsilon$ in both the positive and negative directions. In Figure \ref{fig:color_robustness}, we show the measure of the color extractor on {\sc STOP} images for varying values of $\epsilon$. We observe that the robustness of the color extractor is extremely high for small values of $\epsilon$, and then steadily decreases. Upon closer examination, we find that many of the points the color extractor is non-robust on for small values of $\epsilon$ are points that are very close to a different color boundary, often due to noisy images. We provide a few examples in Figure \ref{fig:nonrobust_examples}. In some cases, like in Figures \ref{fig:bluebad1} and \ref{fig:bluebad2}, the sign has a blueish tint, often due to poor lighting. In other cases, like Figure \ref{fig:yellowbad}, the blurriness hinders correct sign localization (see Section~\ref{sec:signlocalization}). The differences between robustness for blue and yellow for higher $\epsilon$ values is due to a smaller hue distance between red and yellow as compared to between red and blue. For smaller values of $\epsilon$, the difference is due to dataset artifacts -- more {\sc Stop} signs with very poor lighting  in the dataset were closer to having a bluish hue than a yellowish hue (see Figure~\ref{fig:nonrobust_examples} for a few examples).

\section{Additional Theorem}\label{app:theorem}

\begin{theorem} Consider a classifier $F: \mathcal{X} \rightarrow [k]$ and suppose we have access to a group feature extractor $T: \mathcal{X} \rightarrow [m]$ as well as a labels mapping $G: [m] \rightarrow 2^{[k]}$. Consider the augmented classifier $C(x) = F(x) \cap G(T(x))$. If $T(\cdot)$ is robust over $\mathcal{P}$ with respect to $\gamma$, then for all $z \in B(x,\gamma)$, $C(z)$ is non-empty if and only if $F(z) \in G(T(x))$.
\end{theorem}

Above theorem holds because robustness of $T(.)$ implies robustness of $G(T(.))$ from Theorem 2. Thus, the label of $C(.)$ for both $x$ and $z \in B(x, \gamma)$ must be in $G(T(x))$, ruling out targeted attacks that change label of $F(x)$ to a label not in $G(T(x))$. 

As an example scenario of the above theorem, suppose $x \in \mathcal{P}$ is an image of a {\sc stop} sign. $T(x)$ is determined to be red. Then, $G(T(x))$ is the set of sign labels that can be red, e.g., a set including the {\sc Stop} sign and {\sc Do Not Enter} sign.  Let's assume that normal case that $F$ classifies the sign $x$ correctly. Then, $C(x)$ will also give a correct classification. Furthermore, for an arbitrary input $z \in B(x, \gamma)$, since $G(T(z)) = G(T(x))$ due to robustness of $T$, label of $C(z)$ is restricted to be either $\emptyset$ or in the set of red signs, $G(T(x))$. $C(z) = \emptyset$ implies an inconsistency between the two outputs of $F$ and $G(T(\cdot))$ on input $z$, suggesting a problem, which may require human inspection or another intervention to resolve. A non-empty result implies that the two inputs are of the same color, though not necessarily the same label.

\end{document}